\documentclass[cameraready]{Interspeech}

\usepackage{multirow}
\usepackage[most]{tcolorbox}
\usepackage{xcolor}
\usepackage{comment}

\usepackage[activate]{microtype}
\title{XAI-Grounded Explanation Generation for Speech Deepfake Detection with Training-Free Multimodal Large Language Models}

\author{Yupei Li$^{1,2}$, Qiyang Sun$^{1}$, Xiaoliang Wu$^{3}$, Chenxi Wang$^4$, Berrak Sisman$^{5}$, Björn W. Schuller$^{1,2}$}

\address{
\textsuperscript{1} Imperial College London, United Kingdom\\
\textsuperscript{2} Technical University of Munich, Germany\\
\textsuperscript{3} University of Southampton, United Kingdom\\
\textsuperscript{4} MBZUAI, United Arab Emirates\\
\textsuperscript{5} Johns Hopkins University, United States of America
}

\email{yl7622@ic.ac.uk, q.sun23@imperial.ac.uk, xiaoliang.wu@soton.ac.uk, Chenxi.Wang@mbzuai.ac.ae, sisman@jhu.edu, schuller@tum.de}

\keywords{Grounded explanation, Explainable artificial intelligence, Multimodal large language models, Speech deepfake detection, Dataset creation}

\begin{document}

\maketitle

\begin{abstract}
Speech deepfake detection (SDD) systems require trustworthy explanations for reliable decision-making. Existing explanation ways mainly fall into two categories. Traditional explainable AI (XAI), such as gradient-based attribution, produces low-level attribution signals tightly coupled with model decisions, and harder to be understood by human than natural language explanations. Meanwhile, large language model (LLM)-based explanation generation often produces generic and ungrounded descriptions due to the lack of heuristic evidence and task-specific supervision, stemming from limited grounded explanation datasets for SDD. We therefore propose a training-free explanation framework that integrates XAI evidence with multimodal LLMs to generate grounded and specific explanations. Using the PartialSpoof dataset, we construct a grounded explanation dataset and show that methods with XAI increase inside accuracy by over 45\%, verified through human evaluation and faithfulness checks.
\end{abstract}

\section{Introduction}

Speech deepfake detection (SDD) has been extensively investigated using both traditional deep learning models \cite{rana2022deepfake} and large language models (LLMs) \cite{li2025dfallm, gu2025allm4add}. Although often formulated as a binary classification task, SDD remains inherently challenging. This is because it does not merely require determining whether a speech sample is bona fide or spoofed; rather, it necessitates a principled justification of why a given sample is regarded as fake, so as to avoid arbitrary or ungrounded decisions. This explanatory requirement may constitute a fundamental factor underlying the limited generalisability of current SDD systems. Many existing models tend to rely on superficial statistical artefacts specific to particular synthesis techniques such as watermarking, rather than capturing the intrinsic generative mechanisms of spoofed speech \cite{zong2025audiomarknet}. Therefore, the development of trustworthy and interpretable explanations is not merely desirable but essential for achieving reliable, and responsible detection outcomes \cite{cirillo2025explainability}.

Explanations in current approaches are primarily derived from traditional explainable artificial intelligence (XAI) techniques and, more recently, from LLMs that generate natural language rationales. Traditional XAI methods take advantage of mathematically grounded attribution mechanisms, which provide theoretical support for the validity and faithfulness of their outputs. For instance, Integrated Gradients (IG) \cite{sundararajan2017axiomatic} quantify feature importance by attributing prediction changes to input perturbations along a continuous path from a baseline; saliency maps \cite{simonyan2013deep} highlight the most influential input dimensions based on gradient sensitivity; Local Interpretable Model-agnostic Explanations (LIME) \cite{ribeiro2016should} approximate the model locally with an interpretable surrogate to estimate feature contributions; and SHapley Additive exPlanations (SHAP) estimate feature contributions based on Shapley values from cooperative game theory \cite{lundberg2017unified}. However, these methods predominantly rely on locality or linearity assumptions. Moreover, they require access to the original decision-making model, regardless of whether the XAI technique is model-specific or model-agnostic \cite{li2025explainable, sun2025explainable}.

On the other hand, LLM-based explanations can be generated independently of the original detection model, without requiring gradient access or architectural transparency \cite{pang2024generating}. By articulating rationales in natural language rather than low-level spectral visualisations, they are generally more interpretable to human users. However, producing reliable explanations often demands either strong reasoning capabilities, which remain underdeveloped in current audio LLMs \cite{peng2025survey}, or carefully curated training data to guide the model’s attention towards relevant acoustic cues. Without such constraints, LLM-generated explanations are prone to hallucination and may remain superficial, offering only high-level descriptions rather than faithful accounts of the underlying decision process \cite{sahoo2024comprehensive}.

Specifically within the SDD domain, approaches explicitly designed for explainable detection remain limited. Existing studies predominantly rely on traditional XAI techniques \cite{govindu2023deepfake, channing2024toward, akman2025audio}, which have shown effectiveness in providing feature-level attributions. In contrast, LLM-based approaches typically follow a post hoc explanation pipeline similar to that discussed above \cite{xie2026interpretable}, while novelly incorporating multiple LLMs for iterative self-consistency checking \cite{wang2022self}. However, it still lacks principled heuristic mechanisms for assessing explanation quality, and remains an absence of dedicated datasets tailored to SDD explanation. Also, validating the quality of LLM-generated explanations presents an additional challenge \cite{chang2024survey}, as it is hard to objectively evaluate the flexible output with existing metrics to capture faithfulness and factual correctness.

Therefore, to address the limitations of traditional XAI methods, including their dependency on access to the underlying decision model and limited flexibility, as well as the unverifiability and lack of structural support in purely LLM-generated explanations, and to bridge the absence of dedicated explainable SDD datasets, we make the following \textbf{contribution}. First, Unlike prior post-hoc LLM explanation pipelines that rely solely on textual prompts or single-model attribution, our approach introduces cross-model XAI aggregation and fidelity-driven validation to produce semantically rich and specific explanations with less hallucination, while establishing a principled framework for explanation generation and verification. Second, we construct and publicly release a large-scale explainable SDD dataset based on the PartialSpoof dataset \cite{zhang2022partialspoof}, comprising approximately 65,000 explanation instances, to support future research in explainable SDD. \textbf{Codes and data are in \href{https://github.com/glam-imperial/xai-grounded-speech-deepfake}{Github Link} \footnote{\url{https://github.com/glam-imperial/xai-grounded-speech-deepfake}}.}

\section{Methodology: XAI-Grounded Explanation Generation via LLMs (XGEG)}

\begin{figure*}[h]
    \centering
    \includegraphics[width=0.85\linewidth]{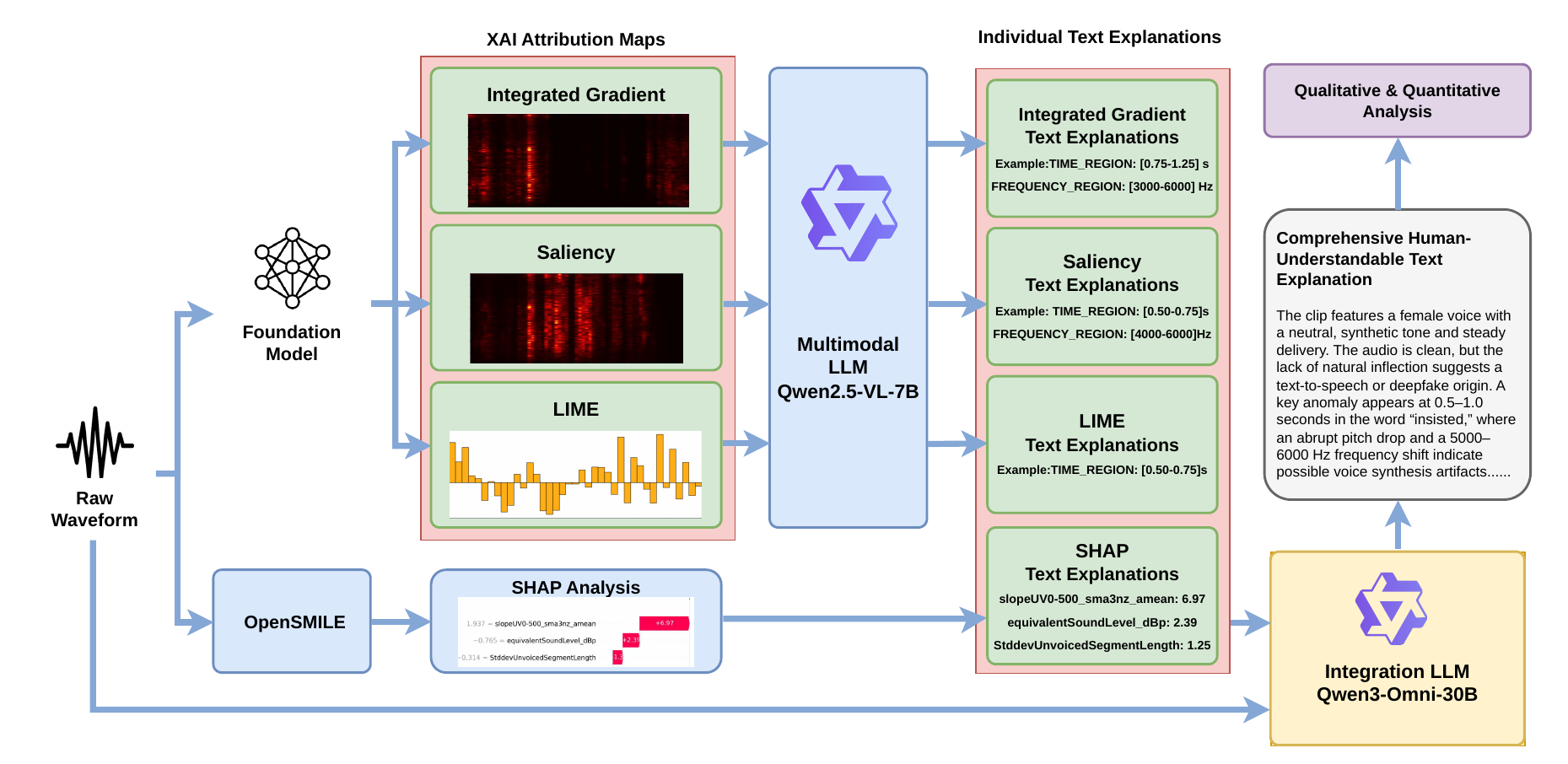}
    \caption{XGEG pipeline. Raw audio is processed by multiple deepfake detection models to generate attribution maps, which are interpreted by a multimodal LLM into notion-level explanations. Meanwhile, acoustic features are extracted with openSMILE, classified by an MLP, and analyzed using SHAP for feature-level attributions. These explanations and attributions are integrated by an LLM to produce the final explanation dataset for evaluation.}
    \label{fig:pipeline}
    \vspace{-0.4cm}
\end{figure*}

As aforementioned, we aim to leverage traditional XAI methods to guide and provide supportive evidence for LLM-based generation, thereby obtaining more trustworthy and specific explanations with less hallucinations. The overall pipeline is illustrated in Figure \ref{fig:pipeline}. Given our objective of proposing a generalizable pipeline, we primarily rely on a training-free framework to enhance scalability across diverse dataset generation settings.

To provide traditional XAI (IG, LIME, and Saliency, selected as representatives) evidence as guidance for LLM-based explanation generation, we employ three pre-trained SDD models based on wav2vec 2.0, HuBERT \cite{antideepfake_2025}, and WavLM \cite{combei2024wavlm} (checkpoint provided) as foundation models. These models are used directly without additional fine-tuning on the target dataset, ensuring an efficient and scalable pipeline without the need to retrain a dedicated detector for each new explanation task. In addition to spectrogram-based evidence, we provide SHAP-derived importance scores for acoustic features extracted using the openSMILE eGeMAPSv02 feature set. Specifically, we train a lightweight four-layer multilayer perceptron (MLP) \cite{rumelhart1986learning} on the training split to estimate feature contributions, deliberately keeping the classifier simple. As our goal is to generate high-quality explanations rather than optimise detection performance, potential data leakage in classification accuracy does not materially affect the validity of the explanatory analysis.

Furthermore, we adopt the widely used PartialSpoof dataset \cite{zhang2022partialspoof} in the work, as it provides precise temporal annotated labels indicating which segments are spoofed. The dataset comprises approximately 25k, 25k, and 71k samples for training, development, and testing, respectively, which are sufficient to support our explanation generation pipeline.

As a sanity check, we evaluate the pre-trained detection models on the selected dataset, reporting Accuracy, F1 and Equal Error Rate (EER) in Table \ref{tab:model_performance}. The results show consistently strong performance across models, suggesting that their predictions are sufficiently reliable to serve as a stable foundation for subsequent XAI analysis. Moreover, when constructing the explainable dataset, we retain only those samples that are correctly classified by all four models. This reduces the risk of erroneous decisions propagating misleading attribution signals and degrading the quality of LLM-generated explanations. We focus on spoofed samples, as explanations for bona fide speech are generally uninformative and typically rely on the absence of anomalies (e.g., ``no acoustic inconsistency detected''). After this filtering process, the dataset comprises around 15k, 15k, and 35k samples for training, development, and testing, respectively.

\begin{table}[t]
\centering
\caption{Performance of Pre-trained Models and MLP Classifier on the PartialSpoof Dataset}
\label{tab:model_performance}
\resizebox{0.9\columnwidth}{!}{
\begin{tabular}{l l c c c}
\hline
\textbf{Model} & \textbf{Split} & \textbf{Accuracy $\uparrow$} & \textbf{F1-score $\uparrow$} & \textbf{EER $\downarrow$} \\
\hline

\multirow{3}{*}{HuBERT} 
& Train & .712 & .414 & .122 \\
& Validation & .697 & .403 & .119 \\
& Test  & .715 & .419 & .108 \\
\hline

\multirow{3}{*}{Wav2Vec 2.0} 
& Train & .709 & .411 & .119 \\
& Validation & .694 & .401 & .109 \\
& Test  & .717 & .421 & .097 \\
\hline

\multirow{3}{*}{WavLM} 
& Train & .703 & .406 & .078 \\
& Validation & .690 & .398 & .067 \\
& Test  & .700 & .408 & .074 \\
\hline

\multirow{3}{*}{MLP} 
& Train & .981 & .912 & .024 \\
& Valid & .961 & .813 & .055 \\
& Test  & .954 & .795 & .063 \\
\hline

\end{tabular}
}
\vspace{-0.7cm}
\end{table}


After obtaining spectrogram-based explanations from IG, LIME, and Saliency for the three pre-trained models, these heatmaps are then provided as input to the vision-LLM Qwen2.5-VL-7B \cite{qwen2.5-VL}, which has shown strong image understanding capabilities. Using a carefully designed prompt (full version provided in the released code), we instruct the model to summarise abnormal regions in terms of their temporal and frequency ranges, the two core dimensions of time–frequency speech representation.

Together with the top three acoustic features ranked by SHAP importance scores, these time–frequency summaries are provided as input to the multimodal LLM Qwen3-Omni-30B \cite{Qwen3-Omni}. This design follows recent research exploring LLM-based explanation generation \cite{xie2026interpretable}. Qwen3-Omni-30B has shown strong performance in audio captioning tasks with comparatively reduced hallucination, making it suitable for synthesising multimodal evidence into coherent textual explanations. Importantly, our objective is not to merely restate the XAI outputs. Instead, we explicitly instruct the model to prioritise the acoustic content of the input and treat the XAI-derived evidence as supporting signals rather than definitive conclusions. Furthermore, the model is prompted to critically analyse the XAI evidence instead of passively reproducing it. To further mitigate hallucination and encourage specificity, we constrain the output to a predefined structured format as follows, ensuring that the generated explanations remain focused, detailed, and evidence-grounded. The full version is provided in the released code. 

\begin{tcolorbox}[
    colback=gray!8,
    colframe=gray!50,
    boxrule=0.4pt,
    arc=2pt,
    left=4pt,
    right=4pt,
    top=4pt,
    bottom=4pt
]

\textbf{I. AUDIO\_ABNORMALITY}\\
TIME\_RANGE: \\
FREQ\_RANGE:

\vspace{2pt}

\textbf{II. EXPLANATION}\\
(Free-text)

\vspace{2pt}

\textbf{III. XAI AGGREGATION}\\
(Indicate XAI contribution and whether it reflects cross-model agreement or single-model evidence.)

\end{tcolorbox}

\section{Results and Discussion}

\subsection{Generated samples}
We adopt the official Hugging Face implementations of all above models and verify their outputs remain consistent across multiple runs. To test whether the XAI signals provide meaningful guidance, rather than being merely paraphrased by the LLM, we design a series of comparative experiments:
\begin{itemize}
    \item Audio-only input, where the integrated LLM receives only the raw audio information without any XAI evidence;
    \item Single-XAI input, where only one attribution method (IG, LIME, or Saliency) derived from a single model (wav2vec2) is provided, resulting in three separate experimental groups;
    \item Full-XAI from one model, where all four XAI signals (IG, LIME, Saliency, and SHAP) from wav2vec 2.0 are supplied;
    \item Cross-model XAI aggregation, where XAI evidence from all three pre-trained models is integrated.
\end{itemize}

We do not conduct a SHAP-only experiment, as SHAP provides feature-level importance without explicit time–frequency localisation, which offers limited guidance to the multimodal LLM in identifying specific abnormal temporal segments. The generated texts as published datasets are in \href{https://github.com/glam-imperial/xai-grounded-speech-deepfake}{Github Link}.

\subsection{Qualitative analysis}
\begin{figure}[htbp]
    \centering
    \begin{minipage}[b]{0.43\linewidth}
        \centering
        \includegraphics[width=\linewidth]{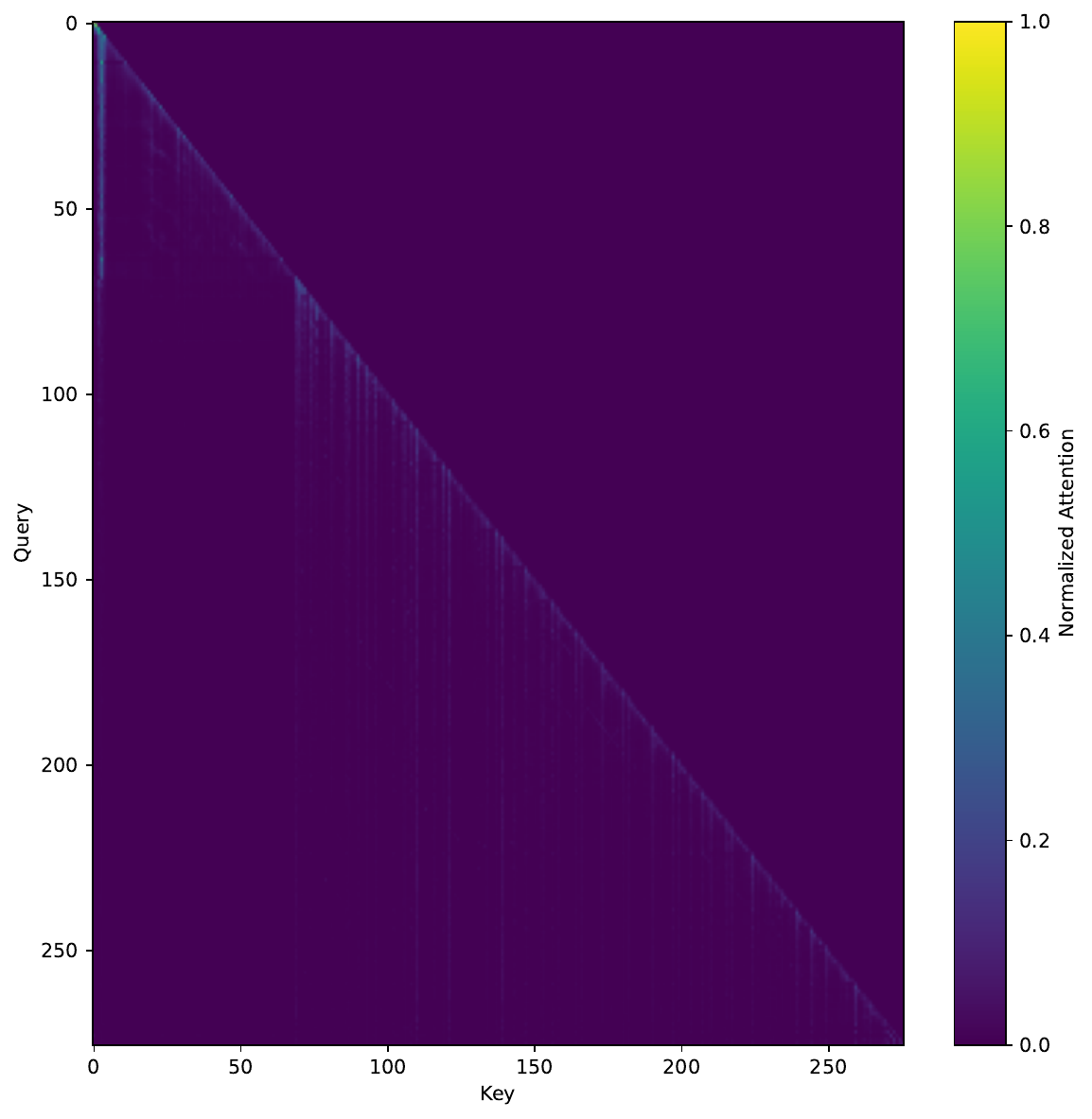}
        \caption{First generation step of the attention map for the first layer on average multiple heads}
        \label{fig:attn1}
    \end{minipage}
    \hspace{0.02 \linewidth} 
    \begin{minipage}[b]{0.43\linewidth}
        \centering
        \includegraphics[width=\linewidth]{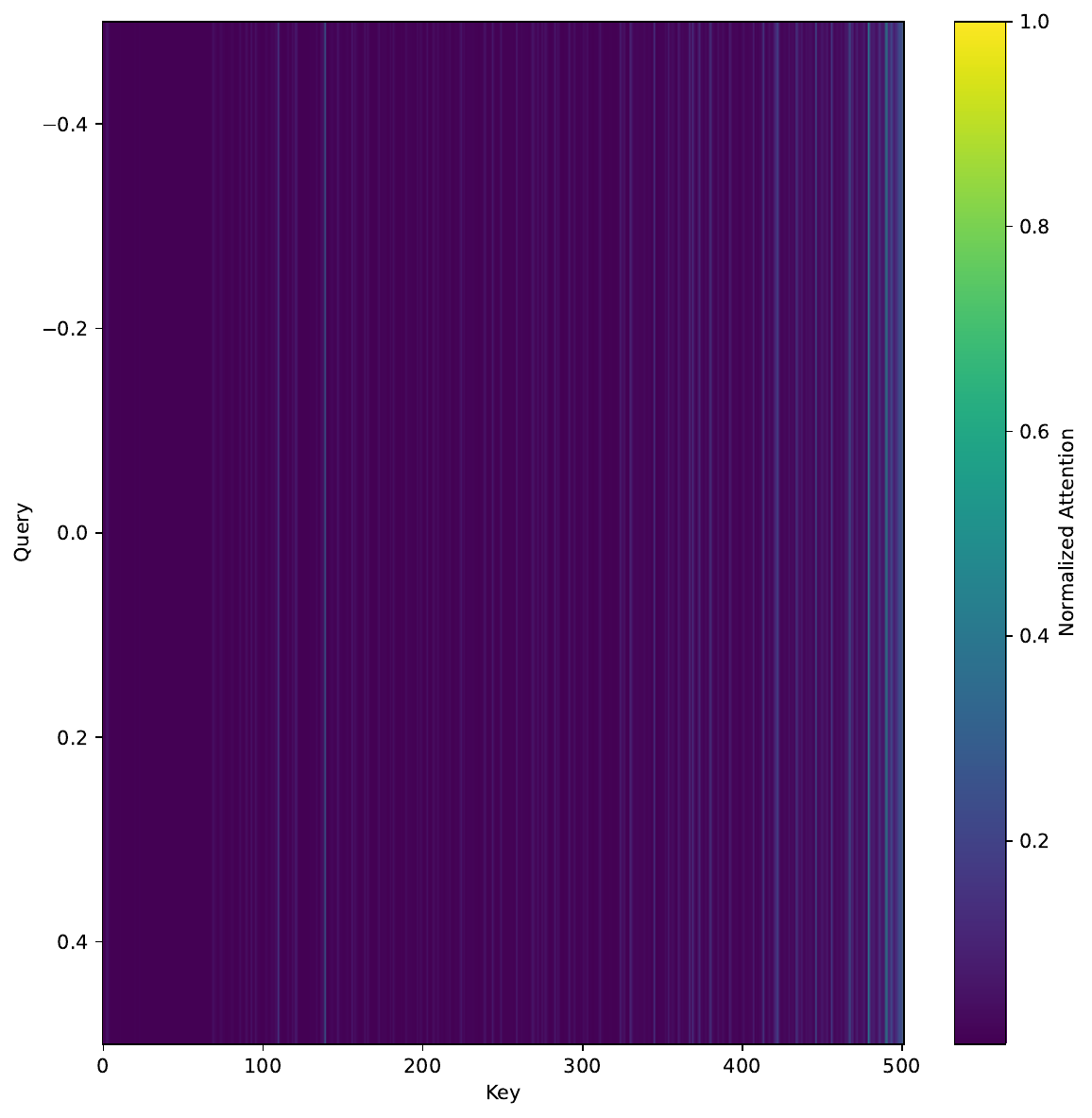}
        \caption{Last generation step of the attention map for the first layer on average multiple heads}
        \label{fig:attn2}
    \end{minipage}
\end{figure}

Qualitative case studies show consistent trends in our model’s effectiveness. Raw audio alone often yields superficial or hallucinated explanations, while lightweight XAI guidance improves precision by identifying temporal and spectral anomalies. The LLM integrates acoustic evidence rather than merely restating XAI outputs, as supported by accurate ASR, precise localisation, and attention visualisations (Figures \ref{fig:attn1}, \ref{fig:attn2}), where audio tokens prepended to the text input receive attention weights and are effectively attended to during decoding. However, reasoning over aggregated XAI signals remains challenging due to LLM inherent reasoning bottlenecks \cite{li2024llms}, which we leave for future work. The model also shows emergent capability \cite{wei2022emergent} in identifying potential deepfake sources such as text-to-speech (TTS) origin. Although systematic evaluation is difficult due to the lack of structured outputs for such abilities, qualitative inspection suggests frequent correctness across many cases. 

\subsection{Human evaluation}
To assess the quality of the generated explanations, we conduct a human evaluation study. A total of 600 explanations corresponding to spoofed audio instances were randomly sampled for human evaluation. Twenty annotators were recruited to assess the explanations. The annotators consisted of 10 males and 10 females, aged between 20 and 30 years. All participants were fluent English users and held at least a bachelor’s degree, with five holding graduate degrees. In addition, five annotators had prior academic or practical experience in audio-related fields. Each annotator is assigned a balanced subset of samples covering all six experimental conditions. In total, each annotator evaluates 30 samples, derived from five distinct original audio clips, with six different explanation formats generated from each clip using six distinct prompts. Participants are informed that all audio samples are spoofed and are instructed to listen carefully to each clip while reading the corresponding explanation. They then rate each explanation on a 5-point Likert scale (1 = lowest, 5 = highest), based solely on its alignment with the audio.


As our goal is to produce explanations understandable to the public, no professional expertise was required. Annotators were recruited without restrictions on academic background, age, or nationality, and were instructed to focus on overall meaning after a brief introduction to the task. The assessment is conducted according to five criteria: \textbf{correctness (C)}, measuring whether the explanation accurately identifies and justifies the abnormal acoustic region; \textbf{evidence support (E)}, evaluating whether the explanation is grounded in observable audio cues rather than hallucinated content; \textbf{specificity (S)}, reflecting the level of detail and precision in the description; \textbf{missing explanation (M)}, assessing whether obvious abnormal regions are omitted; and \textbf{overall preference (O)}, capturing the annotator’s comparative judgement of explanation quality (only selecting 1 sample out of 6). The scores are shown in Table \ref{tab:human_assess}.

\begin{table}[htbp]
\centering
\caption{Human Evaluation Results Under Different Settings}
\resizebox{\columnwidth}{!}{
\begin{tabular}{lccccc}
\toprule
Settings & C$\uparrow$ & E$\uparrow$ & S$\uparrow$ & M$\downarrow$ & O (average number)$\uparrow$ \\
\midrule
Pure Audio (Baseline) & 3.15 & 1.75 & 1.90 & 2.75 & 0.35 \\
\midrule

IG & 3.60 & 3.50 & 3.45 & 2.20 & 0.70 \\
Saliency & 3.75 & 3.65 & 3.30 & \textbf{2.15} & 0.75 \\
LIME & 3.35 & 3.40 & 3.50 & 2.30 & 0.65 \\
All XAI (Single Model) & \textbf{3.90} & \textbf{3.75} & 3.55 & 2.45 & 1.05 \\
All XAI (Three Model) & 3.85 & 3.60 & \textbf{4.30} & 2.30 & \textbf{1.50} \\
\bottomrule
\end{tabular}}
\vspace{-0.4cm}

\label{tab:human_assess}
\end{table}

The results reveal a clear trend indicating that the XAI-guided versions achieve better alignment with human subjective assessments. Notably, the improvements in specificity and evidence support are particularly convincing, suggesting that the XAI-guided models produce fewer hallucinations and provide more detailed justifications. Furthermore, the findings demonstrate that incorporating a greater degree of XAI guidance leads to higher overall recognition. However, to ensure technical accuracy, particularly in aspects such as whether the time period is correctly identified, rigorous quantitative analysis is required.

\subsection{Quantitative analysis}

We therefore conducted a correctness evaluation using a subset of the generated explanations, based on the original development subset of the Partialspoof dataset.

\subsubsection{Intersection over Union (IoU) and Inside Accuracy (IA)}

To evaluate time abnormal period detection, we adopt IoU and IA. IoU is defined as: $IoU = |T_pred \cap T_gt| / |T_pred \cup T_gt|$, where $T_pred$ and $T_gt$ denote the predicted and ground-truth abnormal time intervals, which localisation works often use as a metric \cite{li2025dfallmachievinggeneralizablemultitask, zhang2024deepfake}. IA is defined as the proportion of samples where the predicted time interval is fully contained within the ground-truth interval, which punishes hallucinated guess of the time abnormal period. The results are shown in Table \ref{tab:time_detection}. 

\begin{table}[t]
\centering
\caption{Time period detection performance under different explanation settings.}
\label{tab:time_detection}
\resizebox{0.7\columnwidth}{!}{

\begin{tabular}{lcc}
\hline
\textbf{Settings} & \textbf{IoU} & \textbf{IA} \\
\hline
Pure Audio (Baseline) & \textbf{0.269} & 0.049 \\
\hline

IG & 0.224 & 0.482 \\
Saliency & 0.239 & 0.489 \\
LIME & 0.157 & 0.811 \\
All XAI (Single Model) & 0.242 & \textbf{0.492} \\
All XAI (Three Model) & 0.134 & 0.295 \\
\hline
\end{tabular}}
\vspace{-0.7cm}
\end{table}

Pure Audio achieves the highest IoU due to over-expanded abnormal intervals, but its very low IA indicates poor localization precision. Single XAI methods, such as LIME, often produce unstable and overly narrow regions. Combining multiple XAI sources within one model provides a better balance between IoU and IA, whereas aggregating XAI from multiple models increases reasoning complexity and may reduce stability.
\subsubsection{Area-Normalised Local Logit Sensitivity}

Additionally, we design a set of quantitative attribution experiments as one form of fidelity test \cite{miro2025comprehensive}. Traditional XAI evaluation methods are unsuitable for high-frequency continuous audio tasks. Such methods include direct masking, silencing, or frequency-domain cropping \cite{akman2025improving, li2024detecting}. Aggressive time-frequency masking disrupts phase coherence. This artificially introduces strong digital artefacts. Consequently, this shifts the input out-of-distribution. It causes abnormal fluctuations in the model's confidence for the `Fake' class. It fails to reflect the true causal importance of the extracted features.

To overcome this flaw, we propose the \textbf{Area-Normalised Local Logit Sensitivity} metric. We avoid destructive masking on salient regions. Instead, we apply a minimal multiplicative amplitude perturbation to the target time-frequency area. We set this perturbation to $\epsilon = +1.0\%$ in our experiments. This approach maintains the structural integrity and phase continuity of the audio. We calculate the absolute change in the model's output logit. We then divide this change by the area of the time-frequency region ($\Delta t \times \Delta f$). A higher sensitivity density indicates greater model sensitivity to that specific information.


We compare our multimodal fusion methods with a Pure Audio LLM baseline using localisation regions generated by IG, Saliency, LIME, and 4XAI. As shown in Table \ref{tab:sensitivity_eval}, multimodal methods consistently identify more informative and influential regions than the baseline, particularly with LIME guidance. This may be because LIME relies on a perturbation-based strategy to approximate the model’s local decision boundary, which aligns closely with our analytical framework. The 4XAI variant achieves the best performance, indicating that integrating multiple XAI signals improves localisation of key deepfake speech segments. 
Overall, the XAI (single-model) version achieves the highest quality as the evaluation metrics reported above; therefore, the released dataset follows this configuration.

\begin{table}[h]
\centering
\caption{Quantitative Evaluation of Explainability using Area-Normalised Logit Sensitivity ($\epsilon = +1.0\%$)}
\label{tab:sensitivity_eval}
\resizebox{\columnwidth}{!}{%
\begin{tabular}{lcc}
\toprule
\textbf{Method} & \textbf{Mean Sensitivity Density} ($\times 10^{-6}$) & \textbf{Ratio vs. Baseline} \\
\midrule
Pure Audio (Baseline) & 0.1424 & 1.00$\times$ \\
\midrule
IG & 1.3382 & 9.40$\times$ \\
Saliency & 2.9250 & 20.54$\times$ \\
All XAI (Three Model) & 3.1186 & 21.90$\times$ \\
All XAI (Single Model) & 3.1466 & 22.10$\times$ \\
LIME & \textbf{46.5857} & \textbf{327.20$\times$} \\
\bottomrule
\end{tabular}%
}
\vspace{-0.7cm}

\raggedright
\end{table}
\subsection{Bona fide sample analysis}
We mainly focus on fake audio explanation generation. We also tested some bona fide samples. However, explaining genuine audio is more challenging, as demonstrating the absence of abnormalities is inherently harder than identifying their presence, similar to the difficulty of proving innocence rather than guilt \cite{mumford2021negative}. For bona fide cases, we use a similar prompt, except that the model is not asked to output abnormal time periods, but instead explain why no suspicious evidence is found. Part of one example is: \emph{``The audio sample is a high-fidelity recording of a female speaker saying... The voice is clear, natural, and exhibits subtle, human vocal characteristics such as breathiness...''}.

\section{Conclusion}
In this work, we proposed an XAI aggregation framework that guides training-free LLMs with signals from conventional XAI methods. The framework generates more specific and temporally grounded SDD explanations with reduced hallucination. Experiments, with fidelity analysis and human evaluation, show improved localisation and semantic grounding. By providing interpretable evidence beyond binary decisions, our approach enhances transparency and trust in SDD. Future work will focus on stronger XAI integration and explanation-aware LLM training.
\section{Generative AI Use Disclosure}
We only used Generative AI for grammar check and proofreading of the manuscript.

\bibliographystyle{IEEEtran}
\bibliography{mybib}

\end{document}